# Probabilistic water demand forecasting using quantile regression algorithms

Georgia Papacharalampous[*, #], and Andreas Langousis[*]

Department of Civil Engineering, School of Engineering, University of Patras, University Campus, Rio, 26504, Patras, Greece

**Abstract:** Machine and statistical learning algorithms can be reliably automated and applied at scale. Therefore, they can constitute a considerable asset for designing practical forecasting systems, such as those related to urban water demand. Quantile regression algorithms are statistical and machine learning algorithms that can provide probabilistic forecasts in a straightforward way, and have not been applied so far for urban water demand forecasting. In this work, we aim to fill this gap by automating and extensively comparing several quantile-regression-based practical systems for probabilistic one-day ahead urban water demand forecasting. For designing the practical systems, we use five individual algorithms (i.e., the quantile regression, linear boosting, generalized random forest, gradient boosting machine and quantile regression neural network algorithms), their mean combiner and their median combiner. The comparison is conducted by exploiting a large urban water flow dataset, as well as several types of hydrometeorological time series (which are considered as exogenous predictor variables in the forecasting setting). The results mostly favour the practical systems designed using the linear boosting algorithm, probably due to the presence of trends in the urban water flow time series. The forecasts of the mean and median combiners are also found to be skilful in general terms.

* Correspondence to: papacharalampous.georgia@gmail.com, geopap@upatras.gr (Georgia Papacharalampous); andlag@alum.mit.edu (Andreas Langousis)

# Currently at the Department of Engineering, Roma Tre University, Rome, Italy, and the Department of Water Resources and Environmental Engineering, School of Civil Engineering, National Technical University of Athens, Athens, Greece

ORCID profiles: https://orcid.org/0000-0001-5446-954X (Georgia Papacharalampous); https://orcid.org/0000-0002-0643-2520 (Andreas Langousis)

# 1. Introduction

Urban water demand variables are variables with large practical interest (see e.g., Donkor et al., 2014). Their statistical and distributional features have been extensively studied e.g., by Kossieris and Markopoulos (2018), and Kossieris et al. (2019). These features and their relationships with hydrometeorological variables (examined similarly to those of energy demand variables; see Tyralis et al., 2017) play a key role in the design of urban water demand forecasting methodologies (thoroughly reviewed and classified by Donkor et al., 2014). In fact, most of these latter methodologies naturally (attempt to) consider all the periodicities present in the urban water demand data (i.e., daily, weekly and annual) along with several exogenous predictor (mostly hydrometeorological) variables (that are found or assumed to be informative).

Apart from the above-outlined important considerations and, as automatic time series forecasting (Chatfield, 1988; Taylor and Letham, 2018; Hyndman and Khandakar 2008) is an essential requirement for urban water supply management frameworks, emphasis should also be placed on the forecasting methods' wide applicability in operational conditions (in which only limited human intervention is possible). Statistical and machine learning algorithms (see e.g., Hastie, et al., 2009; James et al., 2013; Alpaydin, 2014; Witten et al, 2017) can be reliably automated and applied at scale (Papacharalampous et al., 2019). Therefore, they are befitting and increasingly adopted for solving urban water demand forecasting problems (see e.g., Herrera et al., 2010, 2011; Quilty et al., 2016; Duerr et al., 2018; Quilty and Adamowski, 2018; Lee and Derrible, 2020; Smolak et al., 2020; Xenochristou and Kapelan, 2020; Xenochristou et al., 2020; Nunes Carvalho et al., 2021; Xenochristou et al., 2021), and several other water informatics problems (see e.g., Tyralis and Papacharalampous, 2017; Althoff et al., 2020a,b, 2021a; Sahoo et al., 2019; Markonis and Strnad, 2020; Rahman et al., 2020a,b; Xu et al., 2020a,b; Tyralis et al., 2021a; Scheuer et al., 2021).

While most of the urban water demand forecasting methodologies are designed to produce mean-value forecasts (see again the review by Donkor et al., 2014), there are also a few recent ones issuing probabilistic forecasts (e.g., Gagliardi et al., 2017). The latter methodologies include some that are based on machine and statistical learning algorithms (e.g., Quilty et al., 2019; Kley-Holsteg and Zie, 2020; Quilty and Adamowski, 2020). However, currently they do not include quantile-regression-based methodologies (see Section 2), although such methodologies and their related concepts have already been exploited in other water informatics contexts, such as contexts focusing on the probabilistic prediction of hydrological signatures (e.g., Tyralis et al, 2021b) or probabilistic hydrological post-processing contexts (e.g., Dogulu et al., 2015; Tyralis et al., 2019a; Papacharalampous et al., 2020; Althoff et al., 2021b).

In this work, we aim to fill this gap by designing the first practical systems for urban water demand forecasting based on the –new for the field– general-purpose concept of quantile regression algorithms. We also aim to provide the first extensive comparison of such practical systems by using one of the largest datasets used so far in the field (see also the datasets in Duerr et al., 2018; Xenochristou and Kapelan, 2020; Xenochristou et al., 2020, 2021). In our experimental investigations, we emphasize at complying with the principles of forecasting (Hyndman and Athanasopoulos, 2018). We additionally consider one of the largest sets of predictor variables used in the field and provide large-scale results on their relative importance.

## 2. Methods and concepts

We design our practical systems based on several –new for the field– concepts. The main of these concepts is the focus on modelling the relationships between the predictors and conditional quantiles of the predictand (i.e., the core concept of quantile regression algorithms) instead of modelling the relationship between the predictors and the conditional mean of the predictand (as done by standard regression algorithms). We exploit this specific concept by applying seven quantile regression algorithms (see Table 1) that are characterized by diverse algorithmic features and exhibit varying performances in real-world problems (with their relative performance being dependent on the problem).

Table 1. Quantile regression algorithms automated and/or applied in this work.

| S/n | Quantile regression algorithm | Indicative application details | Type of algorithm |
|---|---|---|---|
| 1 | Quantile regression | – | Individual algorithm |
| 2 | Linear boosting | {Initial boosting iterations = 2000} | |
| 3 | Generalized random forests | {Number of trees = 2000} | |
| 4 | Gradient boosting machine | {Number of trees = 2000} | |
| 5 | Quantile regression neural networks | {Number of hidden nodes = 1, Hidden layer transfer function = sigmoid} | |
| 6 | Mean combiner of forecasts | – | Combiner of forecasts |
| 7 | Median combiner of forecasts | – | |

Among the selected quantile regression algorithms, the simplest one (that also serves as a benchmark for the others) is the linear-in-parameters quantile regression algorithm (simply referred to as "quantile regression" in the literature and in what follows) by Koenker and Bassett (1978; see also Koenker, 2005). An explanation of this algorithm from a practitioner's point of view can be found in the tutorial by Waldmann (2018). In summary, this algorithm performs minimization of the quantile score (see e.g., Gneiting and Raftery, 2007) averaged over all observations. At level $a \in (0, 1)$, the quantile score imposes a penalty equal to $L(r; x)$ to a predictive quantile $r$, when $x$ materializes according to Equation (1). In this equation, I{·} denotes the indicator function.

$$L(r; x) := (r - x)(I\{x \leq r\} - a) \tag{1}$$

This same minimization is also the objective of three of the remaining selected algorithms. These latter algorithms are the linear boosting algorithm (proposed by Bühlmann and Hothorn, 2007), the gradient boosting machine algorithm (proposed by Friedman, 2001), and the quantile regression neural network algorithm (proposed by Taylor, 2000 and improved by Cannon, 2011).

The boosting algorithms belong to the broader family of ensemble learning methods (Sagi and Rokach, 2018). In summary, the concept behind boosting is the composition of a strong learner (i.e., an algorithm with high predictive ability) by iteratively improving (boosting) weak base learners (i.e., algorithms with low predictive ability). These base learners are linear learners for the linear boosting algorithm and decision trees by Breiman et al. (1984) for the gradient boosting machine algorithm, and are added to the ensemble sequentially. At each iteration, the new base learner is trained to minimize the error of the ensemble (composed by all the previously added base learners). The number of iterations should be large enough to ensure proper fitting and small enough to avoid overfitting. More detailed popularizations of the boosting algorithms can be found in the works by Mayr et al. (2014), and Tyralis and Papacharalampous (2021).

The generalized random forest algorithm is another ensemble learning algorithm for quantile regression, which is formulated as a variant of the original random forest algorithm by Breiman (2001). The latter algorithm is widely applied in water science and water informatics (Tyralis et al., 2019b). It works by averaging an ensemble of trees and uses an additional randomization procedure with respect to bagging by Breiman (1996). With respect to quantile regression forests (another variant of the original random forest algorithm for quantile regression), generalized random forests are theoretically expected to be more suitable for modelling heterogeneities.

As it is indicated by their name, quantile regression neural networks are artificial neural networks especially formulated for quantile regression. More generally, the family of artificial neural network algorithms is perhaps the most popular family of machine learning algorithms in water informatics (see e.g., the review by Maier et al., 2010). The main concept behind this specific family is the extraction of linear combinations of the predictor variables followed by the modelling of the predictand as a non-linear function of these linear combinations.

Apart from the five individual algorithms, we also automate and apply two simple combination methodologies, specifically the mean combiner of the forecasts obtained by the five individual algorithms and the median combiner of the same forecasts. Simple combination methodologies are known to be hard-to-beat in practice in the forecasting field (Lichtendahl et al., 2013; Winkler, 2015), but their advantages are not yet widely applied in water science and water informatics (Papacharalampous and Tyralis, 2020).

## 3. Experimental design

We design and automate practical systems for probabilistic one-day ahead urban water demand forecasting. These practical systems are predominantly based on the methods and concepts outlined in the previous section, while their extension to other horizons and/or temporal scales is possible, and would be a straightforward process from an algorithmic point of view. Information on statistical software is provided in Appendix A.

To extensively compare our practical systems, we use a large urban water flow dataset comprised by recent measurements (taken during 2015–2020) from 54 local automated stations (hereafter referred to as gauges) located at the inlet points of individual district metered areas of the water distribution network of the city of Patras in Western Greece (see Figure 1). The latter exhibit various sizes, topographic and network specific characteristics, as well as data availability. These gauges are part of the "Integrated System for Pressure Management, Remote Operation and Leakage Control of the Water Distribution Network of the City of Patras", which is the largest smart water network (SWN) in Greece, with the Municipal Enterprise of Water Supply and Sewerage of the City of Patras (DEYAP) acting as the competent Authority for its operation and management (Karathanasi and Papageorgakopoulos, 2016).

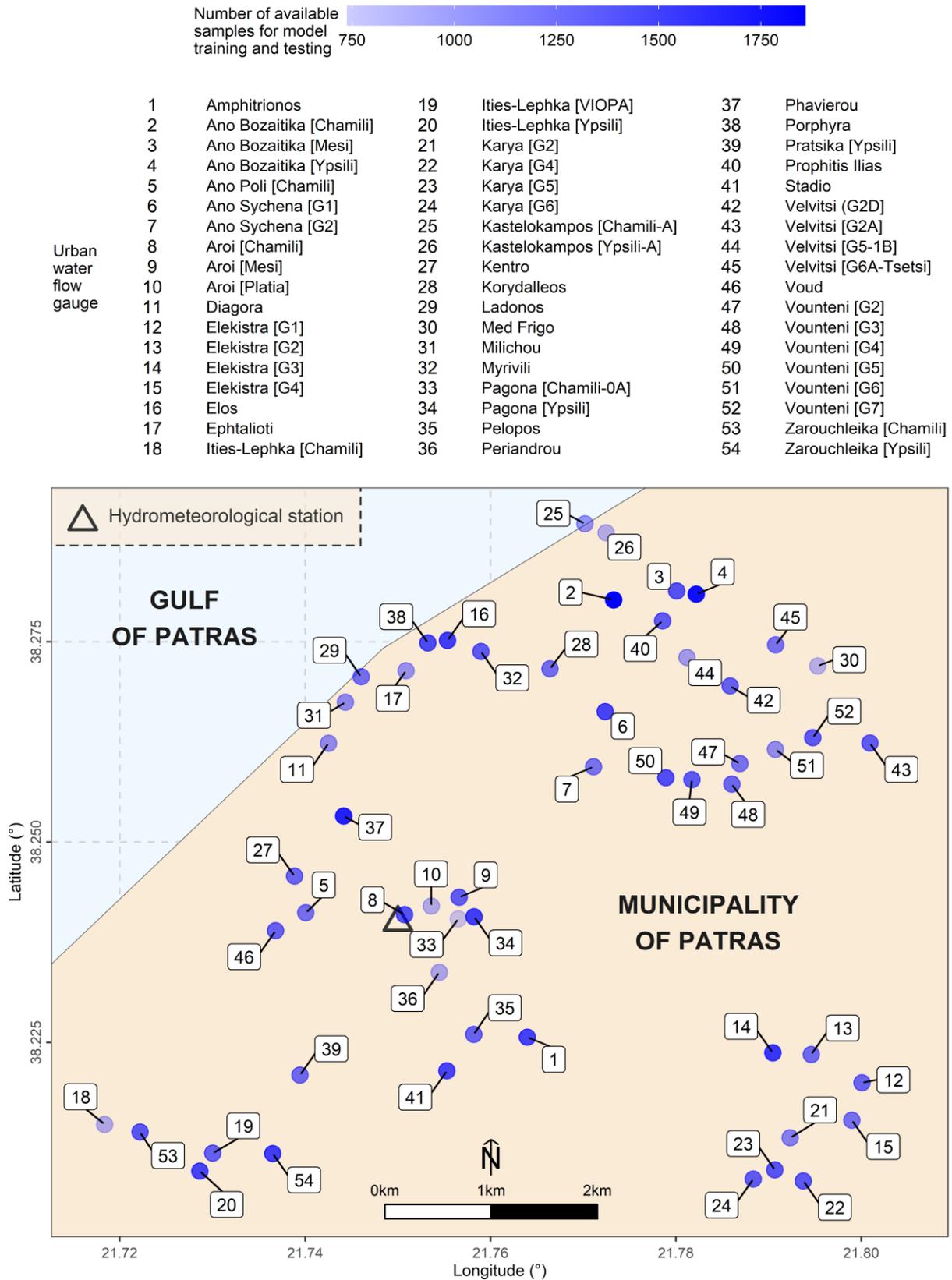

Figure 1. Locations of the 54 urban water flow gauges and the hydrometeorological station exploited in this work, and daily urban water demand data availability.

Starting from the original-measured time series of 1-min resolution, we form time series of daily urban water flow means for each gauge. In our final dataset, we only retain daily values that have resulted from 1-day-long 1-min time series with missing up to the 20% of their values. We further remove the outliers of the daily urban water flow time series, as these outliers have been previously identified with the Friedman's super smoother (Friedman, 1984; Hyndman and Khandakar, 2008), and their unrealistic segments (that might be due to unsuccessful measurements). The means of the final daily urban water flow time series (i.e., those used to form the training and testing samples) range from 0.06 L/s to 95.65 L/s. Their mean is 8.43 L/s and their median 3.12 L/s.

Additionally to the urban water flow time series, we use: high, average and low temperature time series; high, average and low dew point time series; high, average and low humidity time series; high, average and low wind speed time series; high and low atmospheric pressure time series, and total precipitation time series. These time series extend from January 2015 to December 2020 and have no missing values, thereby successfully covering the entire time periods in which urban water demand data are available. Further information on data availability is provided in Appendix B.

Based on the 54 daily average urban water flow time series and the 15 daily hydrometeorological time series, we define and use the following 52 predictors to probabilistically predict average urban water flow at day $t$ ($F_{\text{mean},t}$):

- Average urban water flow at days $\{t-k, k = 1, ..., 7\}$ ($F_{\text{avg},t-k}$)
- High temperature at days $\{t-k, k = 1, ..., 3\}$ ($T_{\text{high},t-k}$)
- Average temperature at days $\{t-k, k = 1, ..., 3\}$ ($T_{\text{avg},t-k}$)
- Low temperature at days $\{t-k, k = 1, ..., 3\}$ ($T_{\text{low},t-k}$)
- High due point at days $\{t-k, k = 1, ..., 3\}$ ($D_{\text{high},t-k}$)
- Average due point at days $\{t-k, k = 1, ..., 3\}$ ($D_{\text{avg},t-k}$)
- Low due point at days $\{t-k, k = 1, ..., 3\}$ ($D_{\text{low},t-k}$)
- High humidity at days $\{t-k, k = 1, ..., 3\}$ ($H_{\text{high},t-k}$)
- Average humidity at days $\{t-k, k = 1, ..., 3\}$ ($H_{\text{avg},t-k}$)
- Low humidity at days $\{t-k, k = 1, ..., 3\}$ ($H_{\text{low},t-k}$)
- High wind speed at days $\{t-k, k = 1, ..., 3\}$ ($S_{\text{high},t-k}$)
- Average wind speed at days $\{t-k, k = 1, ..., 3\}$ ($S_{\text{avg},t-k}$)
- Low wind speed at days $\{t-k, k = 1, ..., 3\}$ ($S_{\text{low},t-k}$)
- High pressure at days $\{t-k, k = 1, ..., 3\}$ ($P_{\text{high},t-k}$)

- Low pressure at days $\{t-k, k = 1, ..., 3\}$ ($P_{\text{low},t-k}$)
- Total precipitation at days $\{t-k, k = 1, ..., 3\}$ ($R_{\text{total},t-k}$)

By considering these predictors, we form the available samples for algorithm training and testing for each gauge. The sizes of these samples are depicted in Figure 1. Apart from using the predictors to design our practical systems, we also assess them –in terms of their relative importance– in solving probabilistic one-day ahead urban water demand forecasting problems for all the gauges. This assessment is made by exploiting the generalized random forest algorithm. Supportively, we examine the Spearman's correlations between the predictand and predictor variables separately for each gauge.

Then, we train each practical system (again separately for each gauge) by using the first 50% of the available samples (e.g., those corresponding to the first two years of a four-year time series), and use them to probabilistically predict $F_{\text{mean},t}$ (by predicting its quantiles at probability levels 2.5%, 10%, 50%, 90% and 97.5%) in the second 50% of the available samples given as inputs the values of the 52 predictors in this same latter half. Finally, we assess the output forecasts by computing the average quantile score according to Equation (1), and summarize the results in terms of (a) relative improvements with respect to the benchmark practical system and (b) rankings.

## 4. Experimental results

### 4.1 Predictor variable importance

As we have used a large number of predictor variables for designing our candidate practical systems, it is meaningful to start by presenting large-scale results on the relative importance of these variables in solving probabilistic one-day ahead urban water demand forecasting problems. We should mention, at this point, that this relative importance cannot be directly inferred by computing correlations (at least not in a straightforward way and partly because of multicollinearity issues). Such correlations are presented in Figure 2 for an arbitrary gauge; however, they can vary significantly from gauge to gauge.

The formal way for inferring this relative importance has, therefore, been adopted and the rankings of the predictor variables have been computed separately for each urban water flow gauge. These rankings are presented in Figure 3, along with their mean values computed over all the urban water flow gauges. The most important predictor variables on average are the lagged urban water flow variables (endogenous predictors) from the most to the least recent ones. The only exception to this latter rule concerns the urban water flow variables observed seven days before the forecast time ($F_{\text{avg},t-7}$). In fact, these variables are found to be more important than the urban water flow variables observed five and six days before the forecast time ($F_{\text{avg},t-5}$ and $F_{\text{avg},t-6}$). This outcome could be attributed to the weekly periodicity.

The second most important category of predictors includes those related to temperature, which together with the endogenous predictors have been found to be the dominant predictor variables almost for all urban water flow gauges. A mixed-variable group of exogenous predictors follows. This group includes average, high and low humidity variables, average and high wind speed variables, high and low atmospheric pressure variables, and average, high and low dew point variables. These predictors have a quite varying ranking from gauge to gauge. On the contrary, the least important predictors are –almost for all urban water flow gauges– the total precipitation and low wind speed ones.

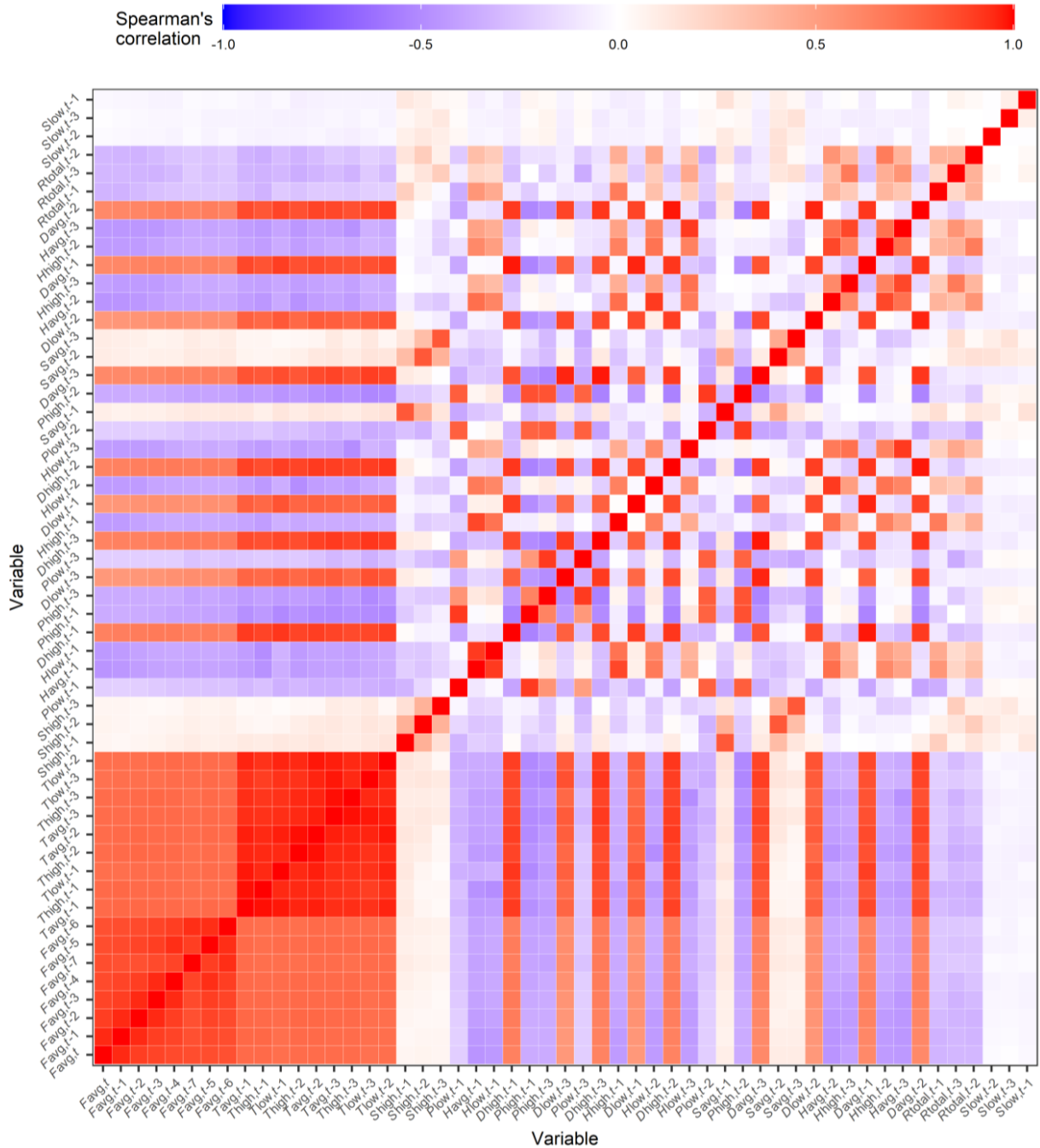

Figure 2. Spearman's correlations between the predictand and predictor variables for an arbitrary gauge. The average rankings of the predictor variables according to their importance in solving probabilistic one-day ahead urban water demand forecasting problems (see Figure 3) have been used for ordering the variables in both axes. The variables are defined in Section 3.

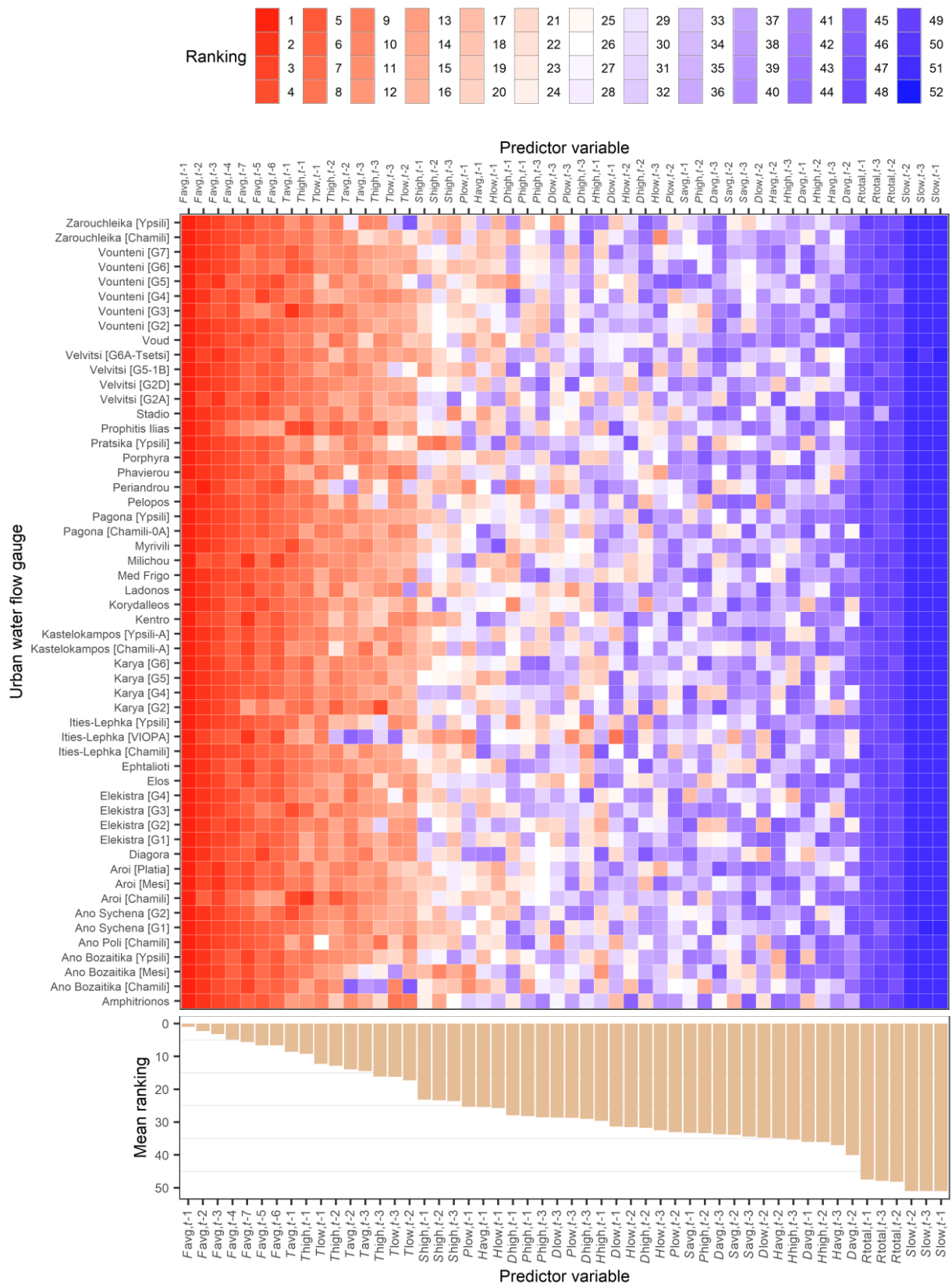

Figure 3. Rankings of the predictor variables according to their importance in solving probabilistic one-day ahead urban water demand forecasting problems using data from the 54 urban water flow gauges. The predictor variables are displayed in the horizontal axis from the most to the least important on average (from the left to the right), as shown in the bottom. The variables are defined in Section 3.

## 4.2 Forecasting performance assessment

As the practical systems of this work have been designed based on seven quantile regression algorithms with different properties, it is meaningful to compare their outputs. A visual inspection of such outputs can be made through Figure 4. This figure presents probabilistic one-day ahead urban water demand forecasts provided by three practical systems for an arbitrary urban water flow time series. We observe that the practical systems utilizing the linear boosting algorithm (Figure 4-a) and the mean combiner of the five individual algorithms (Figure 4-c) produce better prediction intervals than the practical system utilizing the gradient boosting machine algorithm (Figure 4-b). These latter prediction intervals are, in fact, wider than necessary for the largest observed values. Moreover, their width is not as varying as the widths of the prediction intervals obtained by the other two practical systems.

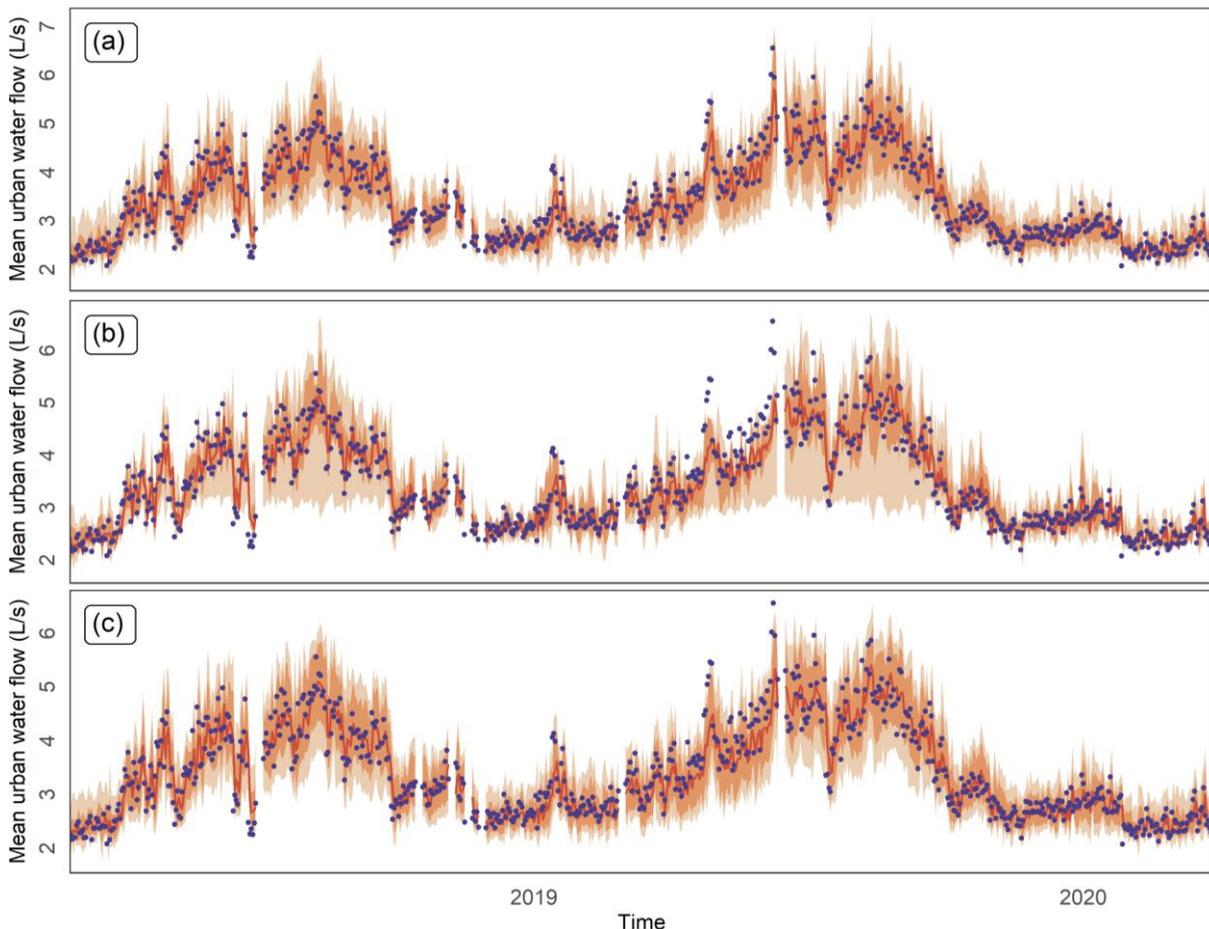

Figure 4. Probabilistic one-day ahead urban water demand forecasts (specifically, the median-value forecasts, 80% prediction intervals and 95% prediction intervals) provided by (a) the linear boosting algorithm, (b) the gradient boosting machine algorithm and (c) the mean combiner of the five individual algorithms (depicted using red nuances) for an arbitrary urban water flow time series (depicted as purple points).

A detailed comparison of the seven algorithms has been conducted regarding their performance when incorporated with the same set of predictors within practical systems for probabilistic one-day ahead urban water demand forecasting. In Figure 5, we present the mean and median values of the average relative improvements provided by the tested algorithms with respect to the quantile regression benchmark in forecasting quantiles of five different levels. Linear boosting has been identified as the best-performing algorithm in terms of both the mean (Figure 5-a) and the median (Figure 5-b) relative improvements, with the mean and median combiners being very close to it (or even better than it) for specific quantile levels. Generalized random forests, gradient boosting machines and quantile regression neural networks perform worse than the linear boosting algorithm and the quantile regression benchmark, probably because of some trends characterizing the urban water demand time series.

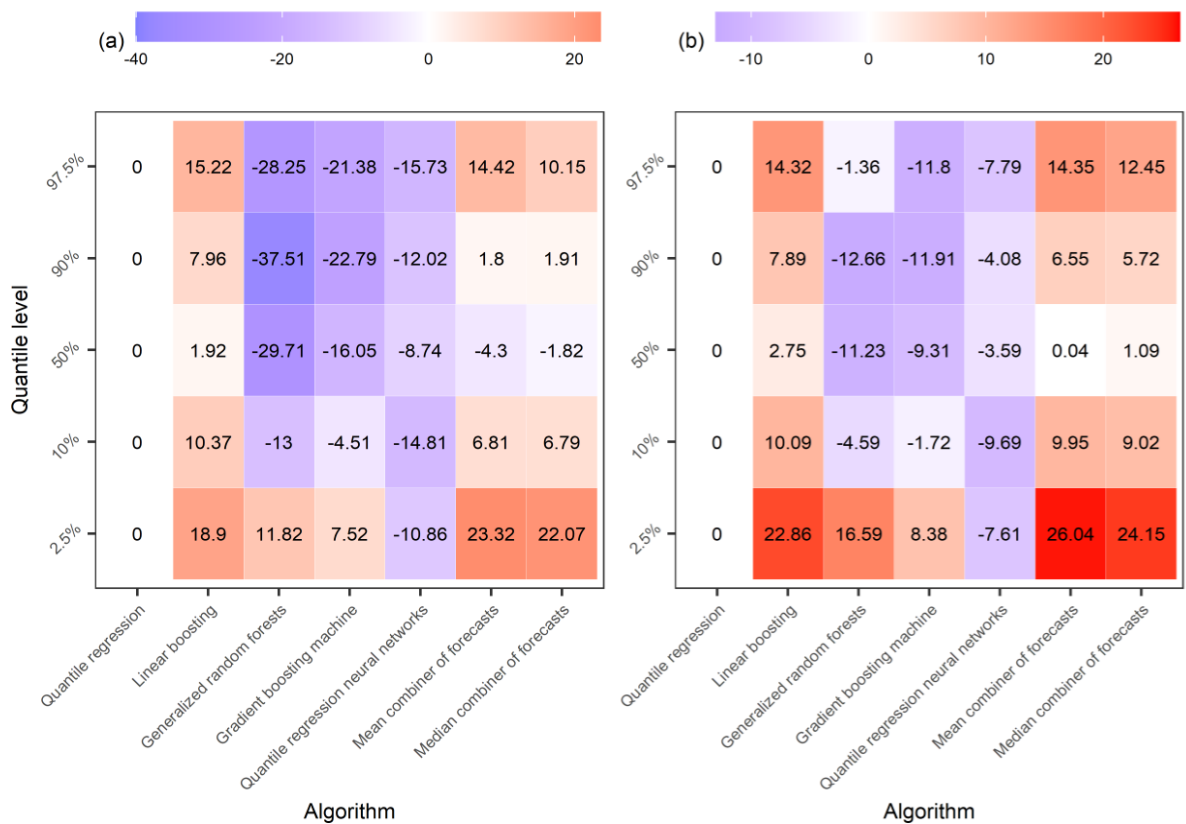

Figure 5. (a) Means and (b) medians of the average relative improvements provided by the tested algorithms with respect to quantile regression (benchmark) in forecasting quantiles of levels 2.5%, 10%, 50%, 90% and 97.5% for the 54 urban water flow gauges.

Lastly, in Figure 6 we present the rankings of the seven algorithms in solving probabilistic one-day ahead forecasting problems as parts of the developed practical systems. These rankings depend to some extent on the quantile level, and mostly favour the linear boosting algorithm and the two combiners of forecasts.

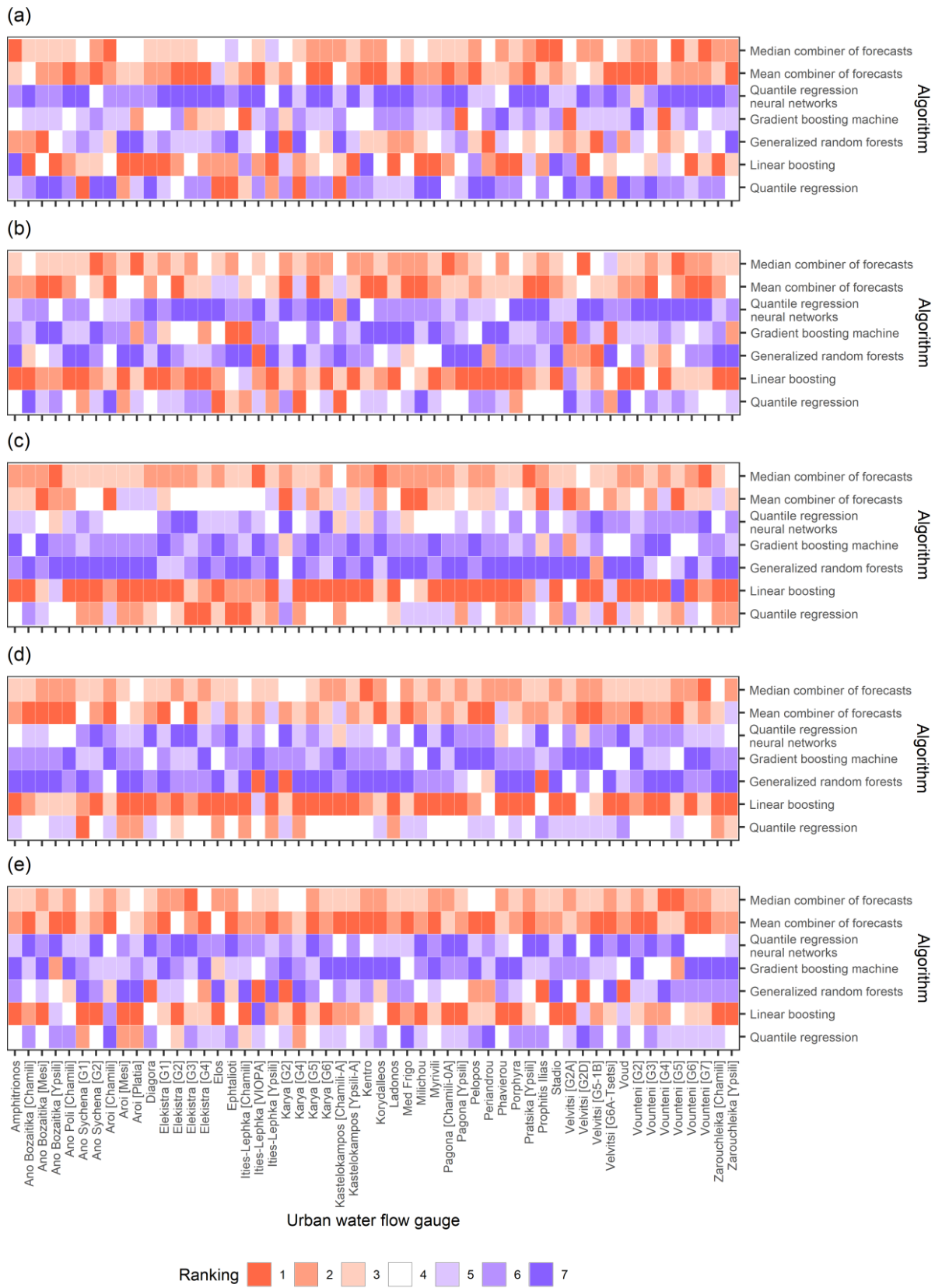

Figure 6. Rankings of the tested algorithms in forecasting quantiles of levels (a) 2.5%, (b) 10%, (c) 50%, (d) 90% and (e) 97.5% for the 54 urban water flow gauges.

## 5. Discussion

Our large-scale results prove the wide applicability of the proposed practical systems. This wide applicability stems from a well-known advantage of machine and statistical learning algorithms, i.e., their appropriateness and efficiency as parts of technical and operational frameworks (Papacharalampous et al., 2019). It also allows their proper assessment and comparison, which relies on big datasets comprising many representative cases (Boulesteix et al., 2018). Herein, we have presented such an assessment. We have also adopted proper scoring rules (see e.g., Gneiting and Raftery, 2007), thereby providing a guide for future studies implementing the general-purpose concept of quantile regression algorithms (see e.g., Waldmann, 2018) for urban water demand forecasting purposes.

Among the most notable findings of this work are the findings on predictor variable importance. We believe that these specific findings could be a good starting point for selecting predictors in similar probabilistic forecasting settings, given also the large scale of our investigations. We have found that the endogenous predictors (i.e., the lagged urban water demand variables) are the most important ones on average, followed by the temperature predictors. Additionally, we have found that precipitation and low wind speed variables are less important than temperature, humidity, wind speed (high and average), atmospheric pressure and dew point variables. Of course, these findings might be –to some extent– dependent on the study area; therefore, they should be confirmed for every case study of interest.

As the boosting and generalized random forest algorithms are not affected by less important predictors by construction (see their property description e.g., in Tyralis et al., 2019b, and Tyralis and Papacharalampous, 2021), we have considered the entire set of endogenous and exogenous predictors herein. Nonetheless, to reduce the computational requirements (which are increased with increasing the number of predictors), one could prefer the use of less predictors even when algorithms with such good properties (in terms of predictors) are selected.

Probably due to the presence of trends in the urban water demand time series, the linear boosting algorithm has been found to be the best-performing quartile regression algorithm in solving probabilistic one-day ahead urban water demand forecasting problems. Boosting algorithms are known for their ability of "garnering wisdom from a council of fools" (Tyralis and Papacharalampous, 2021) and hold their own special place among the machine and statistical learning algorithms. They have also been shown to perform well in probabilistic energy demand forecasting (see the results by Taieb et al., 2016). This might be due to possible similarities characterizing energy and water demand variables (e.g., in terms of trends).

Some last discussions should be made on the good properties of simple combination methodologies. These properties have been already discussed in water science and informatics by Papacharalampous et al. (2020), and include their ability to "harness the wisdom of the crowd" (see e.g., the in-depth discussions in the forecasting field by Lichtendahl et al., 2013; Winkler, 2015). In the present work, these same properties have resulted to forecasts that are –in most cases– comparably skilful to the forecasts provided by the best-performing algorithm. This also implies that the simple combination methodologies might be able to retain their good properties even in cases in which some of the combined forecasts are much less skilful than the rest. This practically means that, by using them instead of using individual algorithms, we can considerably reduce the risk of obtaining a bad forecast at every single forecast attempt. Therefore, they constitute the most reasonable solution to forecasting problems that have not been extensively investigated (by using big datasets and by complying with the forecasting principles and practice) until the time of the forecast.

## 6. Concluding remarks

We have presented a new framework for probabilistic urban water demand forecasting. This framework relies on the general concept of quantile regression algorithms, which has not been investigated before in the field. Inspired by this concept, we have automated and extensively applied practical systems for probabilistic one-day ahead urban water demand forecasting. We have used the forecasts obtained through these practical systems to conduct a large-scale comparison of the incorporated algorithms. The key findings and take-home messages of this work are the following:

- Quantile regression algorithms are straightforward-to-use and, therefore, appropriate for designing practical systems for urban water demand forecasting.
- The linear boosting algorithm has been identified as the best-performing one, when utilized as part of our practical systems.
- The above outcome could be attributed to the presence of trends in the urban water

demand time series.

- For our study area, the temperature predictor variables have been found to be more important in solving probabilistic one-day ahead urban water demand forecasting problems than the remaining exogenous predictor variables.
- Also, humidity, wind speed (high and average), atmospheric pressure and dew point variables have been found to be more important than precipitation and low wind speed variables.
- Simple combination algorithms can reduce the risk of providing bad-quality forecasts.

## Appendix A    Statistical software

The practical systems have been designed and assessed in `R` Programming Language ([R Core Team, 2021](#)). Specifically, we have used the following contributed `R` packages: `data.table` ([Dowle and Srinivasan, 2021](#)), `devtools` ([Wickham et al., 2020b](#)), `dplyr` ([Wickham et al., 2021](#)), `forecast` ([Hyndman and Khandakar, 2008](#); [Hyndman et al., 2020a](#)), `gbm` ([Greenwell et al., 2020](#)), `gdata` ([Warnes et al., 2017](#)), `ggplot2` ([Wickham, 2016a](#); [Wickham et al., 2020a](#)), `ggrepel` ([Slowikowski, 2021](#)), `grf` ([Tibshirani and Athey, 2020](#)), `knitr` ([Xie, 2014](#), [2015](#), [2021](#)), `lubridate` ([Grolemund and Wickham, 2011](#); [Spinu et al., 2020](#)), `maps` ([Brownrigg, et al., 2018](#)), `maptools` ([Bivand and Lewin-Koh, 2021](#)), `mboost` ([Hofner et al., 2014](#); [Hothorn et al., 2020](#)), `qrnn` ([Cannon, 2011](#), [2019](#)), `quantreg` ([Koenker, 2021](#)), `readr` ([Wickham and Hester, 2020](#)), `reshape2` ([Wickham, 2007](#), [2020a](#)), `rmarkdown` ([Xie et al., 2018](#); [Allaire et al., 2021](#)) and `tidyr` ([Wickham, 2020b](#)).

## Appendix B    Data availability

The hydrometeorological data have been retrieved on January 9, 2021 through the following link address: https://www.wunderground.com/dashboard/pws/IU0394U06.

**Declarations of interest:** We declare no conflict of interest.

**Acknowledgements:** We sincerely thank the Municipal Enterprise of Water Supply and Sewerage of Patras (DEYAP) for providing the urban water flow data in their recorded form.

**Funding information:** This research work has been conducted within the project

PerManeNt, which has been co-financed by the European Regional Development Fund of the European Union and Greek national funds through the Operational Program Competitiveness, Entrepreneurship and Innovation, under the call RESEARCH – CREATE – INNOVATE (project code: T2EDK-04177).